\documentclass[12pt]{article}

\usepackage{arabtex}

\usepackage{coling2018}

\usepackage{url}
\usepackage[para,online,flushleft]{threeparttable}
\usepackage{booktabs}
\usepackage{float}
\usepackage{graphicx,subcaption}
\usepackage{smartdiagram}
\usepackage{pgf-pie}
\usepackage{caption} 

\usepackage[stable]{footmisc}

\usepackage{cp1256,utf8}
\setcode{utf8}

\date{}

\begin{document}

\title{Idiom Detection in Sorani Kurdish Texts}


\author{
	\begin{tabular}[t]{c}
		Skala Kamaran Omer and Hossein Hassani\\
		\textnormal{University of Kurdistan Hewl\^er}\\
		\textnormal{Kurdistan Region - Iraq}\\
		{\tt {\{skala.kamaran, hosseinh}\}@ukh.edu.krd}
	\end{tabular}
}

\maketitle

\begin{abstract}
Idiom detection using Natural Language Processing (NLP) is the computerized process of recognizing figurative expressions within a text that convey meanings beyond the literal interpretation of the words. While idiom detection has seen significant progress across various languages, the Kurdish language faces a considerable research gap in this area despite the importance of idioms in tasks like machine translation and sentiment analysis. This study addresses idiom detection in Sorani Kurdish by approaching it as a text classification task using deep learning techniques. To tackle this, we developed a dataset containing 10,580 sentences embedding 101 Sorani Kurdish idioms across diverse contexts. Using this dataset, we developed and evaluated three deep learning models: KuBERT-based transformer sequence classification, a Recurrent Convolutional Neural Network (RCNN), and a BiLSTM model with an attention mechanism. The evaluations revealed that the transformer model, the fine-tuned BERT, consistently outperformed the others, achieving nearly 99\% accuracy while the RCNN achieved 96.5\% and the BiLSTM 80\%. These results highlight the effectiveness of Transformer-based architectures in low-resource languages like Kurdish. This research provides a dataset, three optimized models, and insights into idiom detection, laying a foundation for advancing Kurdish NLP.
\end{abstract}



\section{Introduction}
Idiom detection is an essential area in Natural Language Processing (NLP) due to the challenges idioms pose with their figurative meanings, cultural nuances, and context dependency. Idioms enhance communication by expressing complex ideas succinctly but present difficulties for automated language system. For instance, the Kurdish idiom  ``{\small{\<ئەم ماستە موێکی تێدایە >}}'' (literally means ``this yogurt has a hair in it'') while idiomatically interpreted as ``there is something suspicious about that topic,'' and the idiom ``{\small{\<دار لەسەر بەرد دانانێ>}}'' (literally is translated to ``she/he does not put a stick on a stone'') meaning ``he/she is very lazy''.

 Traditional methods for detecting idioms relied on manual annotations, while modern approaches utilize computational techniques such as machine learning and deep learning. Despite advancements in NLP, idiom detection remains challenging, particularly for low-resource languages like Sorani Kurdish, due to limited annotated datasets and tools. Accurate idiom detection improves applications such as machine translation, sentiment analysis, and literary analysis while preserving cultural heritage.

The study addresses the gap in Kurdish idiom detection by developing and evaluating three models—KuBERT-based transformer sequence classification, RCNN, and BiLSTM with attention—using a manually created dataset of 101 Sorani idioms embedded in over 10,500 sentences. The research objectives include creating a Sorani idiom dataset, training three models, and comparing their performance. Contributions include the creation of a reliable dataset, the development of models for Kurdish idiom detection, and advancing Kurdish NLP research. Limitations involve the dataset's restricted size, which provides a foundation for future work.

The rest of this paper is organized into five sections: overview, related  work, methodology, experimental results, and conclusion with future directions.

\section{An Overview of Idiom Detection}

Idioms present significant challenges in Natural Language Processing (NLP) due to their figurative meanings, which differ from their literal interpretations. These expressions are deeply embedded in cultural and linguistic contexts, making them difficult for automated systems to process. For example, the English idiom ``\emph{kick the bucket}'' means ``to die,'' while the Kurdish idiom ``{{\small{\<بەرد بەبا دەکات>}}'' (``throwing stones to the wind'') conveys futility. Their non-compositional and context-dependent nature complicates their accurate detection by NLP models \cite{sag2002multiword,nunberg1994idioms}. 

In Kurdish, idiom detection is even more complex due to the limited availability of resources and the morphological and syntactic diversity within dialects like Sorani. Sorani idioms range from word-form and phrase-form to sentence-form, where language and cultural nuances are interwoven. For example, ``{{\small{\<سەرو دڵی گرت>}}'' (``he/she took his/her head and heart'') illustrates a sentence-form idiom where metaphor and context are crucial \cite{arif2012idioms,fatih2016linguistics}. Despite advancements in Kurdish NLP, such as the development of resources like KurdNet and parallel corpora, idiomatic expressions remain an underexplored area \cite{amini2021central}. Effective idiom detection is crucial for improving applications such as machine translation and sentiment analysis, as it helps preserve cultural meaning and enhances the accuracy of these systems.

This chapter explores the challenges of idiom detection in low-resource languages like Sorani Kurdish, emphasizing the need for advanced computational models to address these issues. It discusses the various methods and models used in idiom detection, with a focus on their relevance to Kurdish idioms.

\subsection{Technical Background}

The detection of Kurdish idioms in NLP requires specialized models to handle the complexity of language and culture. A primary model used in this study is BERT (Bidirectional Encoder Representations from Transformers), a transformer-based model that captures context from both directions in a sentence, making it effective for understanding the meaning of idiomatic expressions \cite{devlin2018bert}. BERT’s subword tokenization process, which splits words into smaller units, allows it to handle out-of-vocabulary words and languages with complex morphology. Fine-tuning BERT for specific tasks like Kurdish idiom detection helps the model better understand the contextual nuances of idiomatic expressions. For Kurdish, the KuBERT model, a variant of BERT, was developed to handle the language’s rich morphology and dialectal variations, making it particularly well-suited for idiom detection \cite{awlla2025sentiment}.

In addition to BERT, this study employs three key models for idiom detection: a fine-tuned BERT model, a Recurrent Convolutional Neural Network (RCNN), and a hybrid BERT-LSTM model. The fine-tuned BERT model is adapted to identify Kurdish idioms by training it on a specialized dataset of idiomatic expressions. The RCNN combines both recurrent and convolutional layers to capture sequential dependencies and local patterns in the text, making it effective for distinguishing idiomatic expressions from literal ones. The hybrid BERT-LSTM model integrates the contextual understanding of BERT with the sequential modeling power of Long Short-Term Memory (LSTM) networks, enhancing the model’s ability to detect idioms in complex sentence structures.

For tokenization, this study uses three different approaches: the Kurdish BLARK tokenizer, the Kurdish Language Processing Toolkit (KLPT), and the KuBERT tokenizer. The Kurdish BLARK project provides basic resources for Kurdish NLP, including tokenization for both Sorani and Kurmanji dialects \cite{hassani2018blark}. KLPT is a comprehensive toolkit for Sorani Kurdish, offering tokenization and morphological analysis \cite{ahmadi2020klpt}. The KuBERT tokenizer, designed specifically for Kurdish, handles the language’s unique script and morphology \cite{awlla2025sentiment}. These tokenizers ensure accurate text segmentation, which is crucial for detecting idiomatic expressions.

Lastly, the AdamW optimizer, used in the training of these models, helps improve performance by decoupling weight decay from the gradient update process. This regularization technique is particularly important for large models like KuBERT, ensuring stable learning and preventing overfitting. The optimizer's adaptive learning rates facilitate smoother convergence, which is crucial for handling large-scale models and complex NLP tasks.
`

\section{Related Work}

Idiom detection is a key task in NLP, involving the identification of phrases whose meanings cannot be inferred from their constituent words. This challenge is compounded by the structural and usage differences of idioms across languages. Early methods used rule-based systems and statistical models, while more recent approaches leverage supervised and unsupervised machine learning techniques. The latest advancements in idiom detection are driven by deep learning models, particularly transformer-based approaches like BERT, which have set new benchmarks in accuracy. This chapter reviews these evolving methods, covering both traditional and modern techniques for idiom detection.

\subsection{Traditional Approaches}
Traditional idiom detection has employed various machine learning techniques, including rule-based systems, statistical models, and both supervised and unsupervised learning. Rule-based methods, which rely on handcrafted linguistic rules, have shown strong performance but are limited by their reliance on intra-sentential context. For instance, Japanese idiom detection \newcite{hashimoto2006japanese} achieved high accuracy but struggled with non-idiomatic sentences. In Italian, the Lexicon-Grammar framework \newcite{vietri2014lexicon} classified idioms based on syntactic features but faced challenges due to idiomatic ambiguity. Similarly, rule-based systems for Hindi \cite{sinha2014system} and European Portuguese \cite{baptista2015implementing} demonstrated high precision but had limitations in recall. More recently, unsupervised approaches, such as the one proposed by \newcite{verma2015new}, combined web and dictionary knowledge to detect idioms without extensive resources, showing strong performance across multiple datasets. While these traditional approaches have significantly contributed to the field, they remain limited by the need for linguistic resources and their inability to fully capture the contextual nuances of idiomatic expressions.

\subsection{Supervised Traditional Machine Learning Approaches }
Supervised traditional machine learning approaches have significantly improved idiom detection by training algorithms on labeled datasets to identify patterns distinguishing idiomatic from literal expressions. Early work by \newcite{hashimoto2008construction} in Japanese idioms using Word Sense Disambiguation (WSD) achieved high accuracy, and subsequent studies built on this foundation with various classifiers like Support Vector Machines (SVM) and Linear Discriminant Analysis (LDA) to improve detection across different languages \cite{peng2010computing}. More recent methods, such as the use of web dictionaries like Wiktionary \cite{muzny2013automatic}, incorporated new lexical features to enhance detection accuracy, with results showing substantial improvements. Other approaches, such as \newcite{feldman2013automatic}, explored idiom detection through both classification and outlier detection, demonstrating the effectiveness of supervised methods. More recently, \newcite{abebe2023automatic} applied supervised learning to Amharic idioms, achieving high accuracy using SVM and Word2Vec, suggesting further potential through deep learning techniques. These studies highlight the progress of supervised learning in idiom detection, with future work expected to build on these techniques for even better performance.

\subsection{Unsupervised Traditional Machine Learning Approaches }
Unsupervised traditional approaches have significantly contributed to idiom detection by analyzing unlabeled data for patterns and structures. Early work by \newcite{cook2007pulling} focused on the syntactic behavior of idioms, distinguishing idiomatic from literal expressions based on fixedness and flexibility. This approach demonstrated the effectiveness of syntactic cues without labeled data. \newcite{sporleder2009unsupervised} developed a method that assessed idioms' immediate context and cohesive links in discourse, achieving high classification accuracy with an F-score of around 60\%. Similarly, \newcite{fazly2009unsupervised} used statistical measures of lexical and syntactic rigidity, such as Fixednesslex and Fixednesssyn, to identify idioms, even in low-frequency data, enhancing the reliability of detection. These unsupervised methods, which rely on language patterns and structures, provide an effective alternative to supervised approaches, showing that idioms can be identified without labeled data. Traditional approaches, including rule-based, supervised, and unsupervised methods, have made significant progress in idiom detection, offering high accuracy and setting the stage for further research in the field.

\subsection{Word Embeddings, Neural Networks, and Deep Learning Approaches for Idiom Detection}
The use of word embeddings and neural network models, including Word2Vec, GloVe, and context-dependent models like BERT and ELMo, has significantly advanced idiom detection. Early work by \newcite{gharbieh2016word} applied Word2Vec for identifying Verb-Noun idiomatic combinations, achieving high precision (85.5\%-88.3\%) and recall. Subsequent studies, such as \newcite{salton2016idiom} and \newcite{salton2017representations}, leveraged sentence-level embeddings like Skip-Thought Vectors and RNNs with attention mechanisms to enhance idiom detection. \newcite{bai2018text} combined LSTM units with attention to achieve an accuracy of 85.4\%, while \newcite{liu2019bidirectional} introduced BiLSTM-AC, which captured long-range dependencies, outperforming prior models. Transformer-based models like BERT further improved idiom detection by capturing subtle semantic relationships, as shown by \newcite{hashempour2020leveraging}, who compared BERT to other embedding models, and \newcite{kurfali2020disambiguation}, who applied BERT to disambiguate literal and idiomatic meanings. \newcite{nedumpozhimana2021finding} and \newcite{gamage2022bert} demonstrated that BERT-based models significantly improved idiomaticity detection, setting new benchmarks for the field.

These developments illustrate the evolution from early word embedding approaches to sophisticated neural networks and transformer models. Techniques such as BiLSTM, RCNN, and the integration of attention mechanisms have continuously improved idiom detection accuracy. The success of contextualized embeddings like BERT, along with models such as TextCNN and BiLSTM-RNN, has shown that capturing the semantic and syntactic nuances of idioms requires advanced, context-sensitive models. \newcite{liu2019bidirectional}, \newcite{zhou2020multi}, and \newcite{hakkarainen2024automatic} have all highlighted the superior performance of these models in idiom detection tasks, with BERT consistently outperforming traditional methods by better handling multi-word expressions and capturing long-range dependencies within sentences.

\subsection{Summary}
This section reviews idiom detection methods, from traditional machine learning approaches to advanced deep learning techniques. While traditional methods require significant linguistic knowledge, modern models like LSTMs, BiLSTMs, and BERT-based transformers have significantly improved idiom detection by capturing complex linguistic nuances. These models are particularly suited for languages like Kurdish, which have rich morphological structures. The use of KuBERT, a model pre-trained on Sorani Kurdish, holds promise for improving idiom detection in Kurdish texts.

\section{Methodology}
This study focuses on Sorani idioms, a major Kurdish dialect, to ensure the dataset remains homogeneous. Including idioms from other dialects, such as Kurmanji or Hawrami, would introduce variability that could complicate detection efforts. The dataset is compiled from various sources, including books, websites, and agencies. To digitize idioms from books, an OCR tool is utilized, while a web-scraping Python script extracts relevant idioms from websites. Additionally, collaboration with agencies enriches the dataset by incorporating idioms shared specifically for research purposes.

The dataset includes idiomatic expressions from diverse linguistic contexts, covering formal, casual, and culturally specific uses. Idioms are embedded into sentences with various grammatical structures, ensuring the model can recognize them in different forms. To enhance the model's training capabilities, the dataset also includes a class of non-idiomatic sentences, allowing the model to learn the distinction between idiomatic and non-idiomatic expressions. This design ensures efficient training and testing while allowing for easy expansion as additional idioms and contexts are added over time.

Data preprocessing ensures the dataset is clean, consistent, and ready for model training. Initially, the text is normalized and standardized by converting it into a uniform format, eliminating unwanted characters, and correcting spacing. Tokenization is performed using the KuBERT tokenizer, which splits sentences into tokens and encodes them into numeric representations. Attention masks are then created to help the model focus on meaningful tokens while ignoring padding. This preprocessed data is fed into the models, facilitating efficient learning and effectively addressing the complexities of Sorani Kurdish text and idiomatic expressions.

\subsection{Model Architecture}
Idiom detection in this paper is treated as a classification problem due to the figurative nature of idiomatic expressions, which often differ from their literal interpretation. The challenge lies in idioms depending on context, requiring consideration of surrounding words and their relationships to convey the intended meaning. We train three advanced models to address this complexity: KuBERT Transformer, RCNN, and Bi-LSTM with attention. These models are selected for their complementary strengths in handling contextual understanding, sequential dependencies, and selective focus on critical parts of the input. All models use the KuBERT tokenizer, fine-tuned for Kurdish, ensuring consistent tokenization. The AdamW algorithm optimizes each model to prevent overfitting and facilitate fine-tuning. We train each model separately and compare their performance to identify the best model for idiom detection in Kurdish.

\subsubsection{KuBERT-based Transformer Sequence Classification Model}

The KuBERT-based Transformer Sequence Classification Model is a fine-tuned version of BERT, specifically adapted for Sorani Kurdish, utilizing bidirectional encoding to capture contextual information from both preceding and following words. This is essential for idiom detection, where the meaning of expressions often relies on surrounding context. Built on pre-trained Kurdish language representations, the model consists of an embedding layer that converts tokens into dense vectors with positional encodings to maintain sequence order. The core of the model is a 12-layer transformer encoder with multi-head self-attention and feed-forward sub-layers, enabling the model to assess the relevance of words across the sentence and capture the full context for idiom interpretation. The classification head pools outputs from the encoder, often using the [CLS] token, to map sequences to idiomatic categories. Additionally, the model includes a vocabulary expansion to account for Kurdish-specific linguistic features, such as unique expressions and idiomatic phrases. With approximately 110 million parameters and a sequence length of 128 tokens, KuBERT is optimized using the AdamW optimizer and a linear decay learning rate schedule, ensuring strong performance for idiom detection in Sorani Kurdish.

\subsubsection{RCNN Model}
The RCNN model architecture combines the strengths of bidirectional LSTM layers for capturing long-range dependencies and convolutional layers for extracting localized features, making it well-suited for idiom detection tasks. The embedding layer converts input sequences into 128-dimensional dense vectors, which are then processed by a bidirectional LSTM layer with 256 hidden units per direction, capturing context from both past and future words. The LSTM outputs are fed into a convolutional layer with filters of size 3, which extracts n-grams and key phrases, followed by max-pooling to retain the most significant features. The pooled features are passed through a fully connected layer, producing logits for classification, with dropout regularization applied to prevent overfitting. This architecture balances broad contextual understanding and precise pattern recognition, crucial for detecting the figurative meanings of idiomatic expressions. It employs AdamW optimization, cross-entropy loss, and a custom vocabulary enriched with Kurdish-specific tokens for improved performance in Sorani Kurdish idiom detection.

\subsubsection{BiLSTM with Attention Model}
The BiLSTM with Attention model combines fine-tuned BERT embeddings, bidirectional LSTM (BiLSTM) layers, and an attention mechanism for idiom detection in Sorani Kurdish. BERT generates 768-dimensional contextual embeddings, capturing the language's nuances, which are then processed by a BiLSTM layer with 256 hidden units in each direction, learning dependencies in both forward and backward contexts. The attention mechanism highlights the most relevant parts of the sequence, improving classification accuracy. A fully connected layer maps the output to the classification task, while dropout regularization (0.3 rate) prevents overfitting. With around 50 million parameters and a custom Kurdish vocabulary, the model balances complexity and generalization, making it effective for detecting idioms in Kurdish texts.

\subsection{Training and Validation}

The models are implemented in Python 3.8 using the PyCharm IDE, with PyTorch employed for model development to leverage its powerful capabilities for handling layers and optimization processes. Tokenization is performed using the Hugging Face library, which ensures efficient and accurate preparation of input text, while pandas and numpy are utilized for comprehensive data processing. To enhance efficiency and maintain clarity in tracking operations, custom scripts are developed to streamline logging and output handling. For optimization, the models are fine-tuned using the AdamW optimizer, a technique known for its effectiveness in regularizing weights and preventing overfitting. A learning rate of 2e-5 is chosen, which is a standard setting for fine-tuning large pre-trained models and ensures a balanced learning process across all models. 

Building on this robust setup, stratified k-fold cross-validation is applied to the preprocessed data to address class imbalance challenges, ensuring that each fold reflects the idiom class distribution accurately. This approach is particularly significant in idiom detection tasks, as it maintains consistency between training and validation splits, leading to reliable performance metrics and improved generalization across diverse datasets. 

To evaluate the models effectively, k-fold cross-validation is again employed, with performance metrics averaged over multiple folds. This averaging reduces the impact of anomalies, providing robust and reliable measures of the models' capabilities. The evaluation metrics include accuracy, which offers a general sense of performance by measuring the ratio of correct predictions to total predictions, although it may not fully capture performance in imbalanced datasets. Precision, on the other hand, is used to determine the proportion of true positive predictions out of all positive predictions, which is crucial in minimizing false positives during idiom detection. Recall, or sensitivity, measures the proportion of true positive predictions out of all actual positives, ensuring that idioms are not overlooked, even if some non-idioms are misclassified. Finally, the F1-score, which represents the harmonic mean of precision and recall, is used to provide a balanced evaluation by considering both false positives and false negatives, making it especially valuable in scenarios involving imbalanced datasets.

\section{Experiments}

\subsection{ Dataset Design and Size}

The dataset prioritizes depth over breadth, focusing on 101 idioms, each represented by over 100 examples. The example sentences are systematically generated in illustrating 35 different contexts for each idiom, scenarios that involve either formal speech, casual conversation, or certain cultural settings. These 35 contexts are further repeated three times to increase the richness and variation of the dataset with different grammatical structures. For example, an idiom would be in a simple declarative sentence, another in a question, and another in a conditional. Table \ref{table:dataset_structure} depicts the structure of this dataset, where the column idiom\_y is the idiom, X is an example sentence containing the idiom, and Y is the classification number. This design ensures diversity and richness, resulting in a total of 10,581 sentences, which details the idioms, example sentences, and classification labels. A non-idiom class is included to balance the dataset, allowing the model to distinguish idiomatic from non-idiomatic text and avoid overgeneralization, making it applicable in real-world scenarios.

\begin{table}[!ht]
\caption{Example Dataset Showing Kurdish Sorani Idioms with Corresponding Sentence Examples.\label{table:dataset_structure}}
\begin{threeparttable}
\begin{tabular*}{\columnwidth}{@{\extracolsep\fill}p{1cm}p{8cm}p{3cm}@{\extracolsep\fill}}
\toprule
\textbf{y} & \textbf{x} & \textbf{idiom\_y} \\
\midrule
4 & {\small \RL{مەبەستم نەبوو برینت بکولێنمەوە کاتێک باسی سەرکەوتنی براکەمم کرد.}} & {\small \RL{برین کولاندنەوە}} \\
4 & {\small \RL{کاتێک باسی سەرکەوتنی براکەمم کرد، مەبەستم نەبوو برینەکەت بکولێنمەوە.}} & {\small \RL{برین کولاندنەوە}} \\
4 & {\small \RL{برینەکەت کولایەوە کاتێک باسی سەرکەوتنی براکەمم کرد، بەڵام مەبەستم نەبوو.}} & {\small \RL{برین کولاندنەوە}} \\
4 & {\small \RL{شیلان هەستی کرد برینەکەی دەکولێتەوە هەر جارێک کە خوشکەکەی باسی منداڵەکانی دەکات، چونکە خۆی ناتوانێت منداڵی ببێت.}} & {\small \RL{برین کولاندنەوە}} \\
4 & {\small \RL{برینی شیلان دەکولێتەوە هەر کاتێک خوشکەکەی باسی منداڵەکانی دەکات، چونکە خۆی ناتوانێت منداڵی ببێت.}} & {\small \RL{برین کولاندنەوە}} \\
4 & {\small \RL{شیلان ناتوانێت منداڵی ببێت و خوشکەکەی برینەکەی دەکولێنێتەوە هەر کاتێک باسی منداڵەکانی خۆی دەکات.}} & {\small \RL{برین کولاندنەوە}} \\
\bottomrule
\end{tabular*}
\end{threeparttable}
\end{table}

The limited set of idioms with extensive examples allows the model to capture contextual nuances instead of surface patterns. This approach aligns with studies in under-resourced languages, such as \newcite{briskilal2022ensemble}, who used 1,470 idioms with BERT and RoBERTa, and  \newcite{abebe2023automatic}, who worked with 1,000 idioms for Amharic, demonstrating that smaller datasets can achieve significant results. While expanding the idiom set might improve performance, this focused dataset facilitates manageable experimentation with resource-intensive models like BERT.

Constructing the dataset presented challenges, particularly in ensuring high-quality, contextually accurate examples. Students from Salahaddin University, guided by Kurdish linguists, created examples that varied in structure and avoided direct translations to prevent misleading the model. A thorough review by a Kurdish linguist ensured linguistic and contextual reliability. Balancing the dataset with non-idiomatic examples, designed to match idiomatic sentences in form and length, was crucial to prevent bias and enable generalization to new data.

This dataset supports multi-class classification, enabling the detection of specific idioms and distinguishing idiomatic from non-idiomatic text. Such granularity facilitates applications like translation, sentiment analysis, and educational tools. By leveraging BERT’s contextual capabilities and focusing on a rich yet manageable dataset, this study provides a strong foundation for idiom detection in Sorani Kurdish. As the first study of its kind for this language, it highlights the potential for future expansions and refinements, paving the way for advancements in under-resourced language processing.

\subsection{Preprocessed Data}
Data preprocessing is a crucial step in preparing the dataset for modeling, especially for low-resource languages like Sorani Kurdish, where linguistic nuances must be preserved. This section outlines the main steps in preprocessing, including normalization, tokenization with the KuBERT tokenizer, handling of special tokens, and integration with the models.

\subsubsection{Normalization and Standardization}
The first step involves normalizing the dataset to a standard format. This includes transforming input into string format, applying uniform casing, and handling special characters and whitespace. These actions ensure consistency in the data, which is essential for proper tokenization and model performance, particularly when working with languages like Kurdish that have distinct linguistic features.

\subsubsection{Tokenization}
Tokenization is performed using the KuBERT tokenizer, specifically designed for Kurdish text. It divides sentences into subwords, maintaining the language's unique structure. This tokenizer captures the nuances of Sorani Kurdish, including compound words and complex word orders. It produces numerical representations of text by referencing a pre-defined vocabulary tailored to the Kurdish language, which enables the model to interpret idiomatic expressions correctly.

The tokenizer addresses the challenges posed by Kurdish’s morphological richness, handling compound words and word-order variations that might not be well-represented by tokenizers trained on more common languages. KuBERT's pre-training on diverse Kurdish text—spanning domains such as news articles, literature, and social media—enhances its ability to process idiomatic expressions, which are highly context-dependent. Each token is assigned an ID from the model's vocabulary, and special tokens, such as [CLS] and [SEP], are added for classification tasks. The attention mask is also generated to differentiate meaningful tokens from padding, ensuring efficient model processing. Figure \ref{fig:Tokenztion_output} illustrates a tokenized senternce.

\begin{figure}[ht!]
    \centering
    \includegraphics[width=0.6\linewidth]{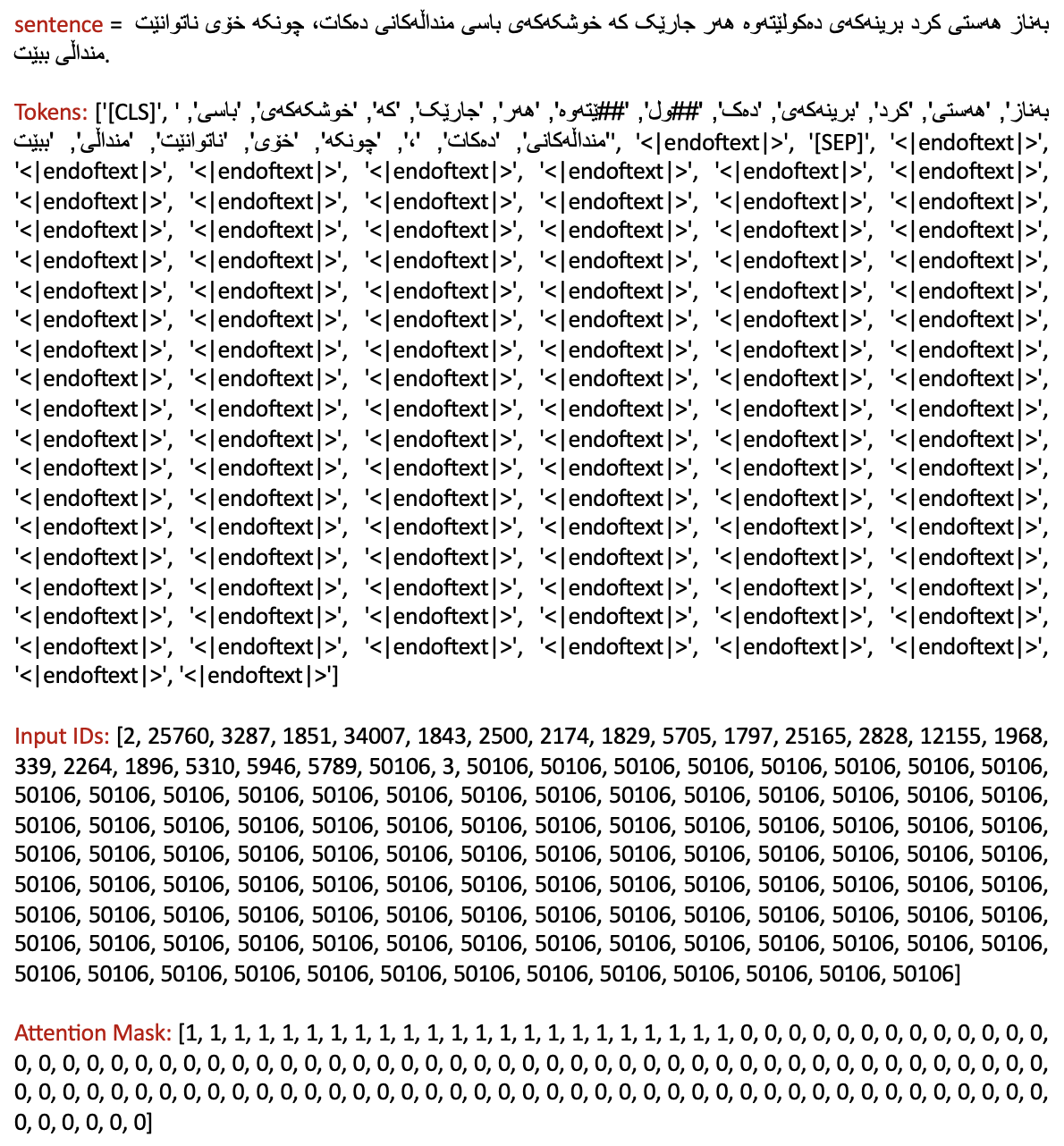}
    \caption{Tokenized Example using KuBERT}
    \label{fig:Tokenztion_output}
\end{figure}

The tokenization process is illustrated in Figure \ref{fig:Tokenztion_output}, which show how sentences are tokenized, padded, and truncated to fit the model's input size requirements.

\subsubsection{Integration with Models}
Once tokenization is complete, the preprocessed data—consisting of input IDs, attention masks, and special tokens—are fed into the model for training and evaluation. The attention mask guides the model to focus on meaningful tokens while ignoring padding. This ensures that the model processes relevant parts of the text, which is crucial for tasks like idiom detection where context plays a significant role.

In summary, preprocessing prepares the data in a structured format that enables the model to effectively learn from the linguistic subtleties of Sorani Kurdish, particularly in the context of idiomatic expressions. Through careful tokenization and the use of specialized tokens, the model can better understand and interpret the complexities of Kurdish language.

\subsection{Training and Validation}

\subsubsection{Environment Setup} 

The models were developed and trained using Python 3.8 within the PyCharm IDE, leveraging the PyTorch framework for efficient implementation. Key libraries included \texttt{torch} for defining model layers and optimization, \texttt{sklearn} for cross-validation and metrics evaluation, and Hugging Face's \texttt{transformers} for tokenization of BERT-based models. Data preprocessing was handled with \texttt{pandas} and \texttt{numpy}, ensuring efficient dataset management. Custom utility scripts were employed for logging and output management, streamlining the workflow. The environment setup prioritized compatibility among tools and libraries, providing a controlled and organized platform for model development.

\subsubsection{Hardware Setup} 

Model training and testing were conducted on a high-performance machine running Windows 11 Pro (64-bit, Version 10.0, Build 22000), powered by an Intel(R) Core(TM) i7-10700KF CPU @ 3.80GHz with 16 logical processors, enabling efficient parallel computations for tasks like backpropagation and hyperparameter tuning. The system featured 32 GB of RAM for handling large datasets and complex model parameters, and a Samsung SSD 870 QVO 1TB (704.5 GB free) for high-speed storage. An NVIDIA GeForce RTX 4060 GPU with 8 GB of dedicated memory accelerated deep-learning tasks, such as training KuBERT Transformer, RCNN, and BiLSTM with Attention models. DirectX 12, including features like Direct3D and AGP Texture Acceleration, further enhanced performance, ensuring efficient and reliable model training and evaluation.

\subsubsection{Data Splitting} 
Nested stratified 5-fold cross-validation was used, with each fold serving as a test set once while the remaining folds were used for training and validation. The training data (80\%) was further split into training (80\%) and validation (20\%) sets for hyperparameter tuning, ensuring robust evaluation and generalization.

\subsubsection{Parameter Settings} 
The hyperparameters for each model were optimized to achieve the best performance. The key settings are summarized in Table~\ref{table:hyperparameter_comparison}.

\begin{table}[!ht]
\caption{Model Hyperparameters\label{table:hyperparameter_comparison}}
\begin{threeparttable}
\begin{tabular*}{\columnwidth}{@{\extracolsep\fill}p{4.5cm}p{3cm}p{3cm}p{2.7cm}@{\extracolsep\fill}}
\toprule
\textbf{Hyperparameters} & \textbf{KuBERT Transformer} & \textbf{BiLSTM with Attention} & \textbf{RCNN} \\
\midrule
Learning Rate & 2e-5 & 2e-5 & 2e-5 \\
Batch Size & 16 & 16 & 16 \\
Number of Epochs & 15 & 50 & 50 \\
Tokenizer & KuBERT & KuBERT & KuBERT \\
Hidden Size & NA & 256  & 256  \\
Embedding Dimension & NA & NA & 128 \\
Dropout Rate & NA & 0.3 & 0.5 \\
\bottomrule
\end{tabular*}
\end{threeparttable}
\end{table}

\subsubsection{Optimization} 
The AdamW optimizer was employed with a learning rate of 2e-5 for all models. KuBERT and BiLSTM used a linear learning rate decay schedule with warm-up steps, ensuring stability during initial training and fine-tuning. The RCNN model was trained with a fixed learning rate, simplifying the optimization process due to its architecture's independence from pre-trained embeddings.

\subsection {Evaluation}
The models were evaluated using a 5-fold cross-validation strategy to ensure robustness and generalization. The dataset was divided into five subsets, with each fold serving as the validation/test set once, while the remaining four were used for training. The evaluation metrics included accuracy, precision, recall, and F1-score, chosen for their relevance to idiom detection. Precision minimized false positives, recall captured true idioms, and F1-score balanced both metrics, especially for the imbalanced dataset. Final metrics, averaged across all folds, provide a realistic measure of the models' effectiveness in real-world idiom detection tasks. Detailed results are discussed in subsequent sections.

\section{Results and Discussions}
The results section presents the outcomes of training the three models—KuBERT Transformer, RCNN, and BiLSTM with Attention—on the same dataset comprising 101 idioms embedded in 10,581 sentences. Using a consistent 5-fold cross-validation approach, with each test set constituting 20\% of the dataset, the models were evaluated using accuracy, precision, recall, and F1-score to ensure a fair comparison. The performance of each model is detailed sequentially: KuBERT Transformer, followed by RCNN, and finally BiLSTM with Attention. Each subsection describes the respective model, its configuration, cross-validation results, and key insights. The section concludes with a comparative analysis of the three models, highlighting their strengths and weaknesses in idiom detection.

\subsection{KuBERT Transformer Model's Results and Performance}
The KuBERT Transformer was trained separately on each of the five folds for 15 epochs with a batch size of 16 and a fixed learning rate of 2e-5. This configuration was determined after preliminary experimentation to ensure stable convergence and effective learning. Validation performance was monitored at regular intervals (every 5 epochs) during training to fine-tune hyperparameters and avoid overfitting. The iterative adjustment process prioritized generalizability across diverse contexts and idiomatic usages.

\paragraph{Comparison of the Folds}  
Figure \ref{fig:Val_Loss} illustrates the validation loss learning curves across all five folds over 15 epochs. The loss decreases significantly during the initial epochs, stabilizing near 0.0 by epoch 5. This consistency across folds reflects a well-tuned model with minimal performance variability, indicating that optimal performance is achieved early. Metrics peak in Fold 3, with accuracy and F1-score reaching approximately 99.4\%, while minor declines in Folds 4 and 5 suggest potential data variance or slight overfitting. Overall, the model consistently performs within 98.5\%-99.5\% across all metrics, showcasing its robustness.

\begin{figure} [ht!]
    \centering
    \includegraphics[width=0.65\linewidth]{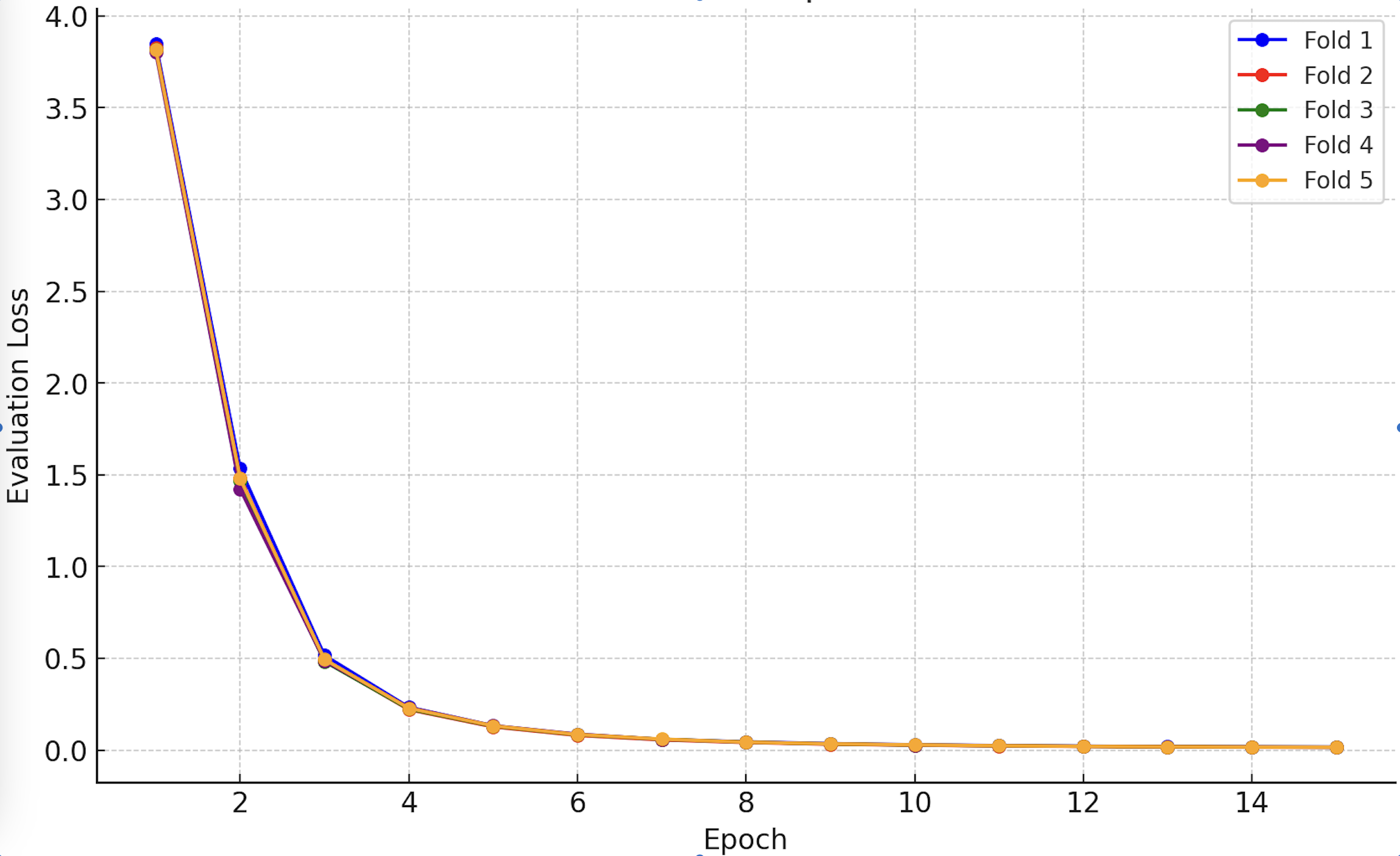}
    \caption{Validation Loss Across All Folds}
    \label{fig:Val_Loss}
\end{figure}

\subsubsection{Overall Performance}  
Averaging results across all folds, the model achieved an overall accuracy of 99.07\%, with precision, recall, and F1-scores of 99.09\%, 99.07\%, and 99.05\%, respectively as shown in Table \ref{table:Transformer_Overall_Performance}. The balanced precision and recall indicate the model's effectiveness in identifying true positives while minimizing false positives. The F1-score highlights the model's robustness and reliability in handling idiomatic expressions across diverse contexts. Training across all folds required 2607.23 minutes (approximately 43.45 hours), emphasizing the computational resources invested to achieve high accuracy.

\begin{table}[!ht]
\caption{Overall Test Performance Summary Across All Folds}
\label{table:Transformer_Overall_Performance}
\begin{threeparttable}
\begin{tabular*}{\columnwidth}{@{\extracolsep\fill}lcccc@{\extracolsep\fill}}
\toprule
\textbf{Metric} & \textbf{Accuracy (\%)} & \textbf{Precision (\%)} & \textbf{Recall (\%)} & \textbf{F1-Score (\%)} \\
\midrule
  Fold 1 & 98.82 & 98.84 & 98.82 & 98.81 \\
Fold 2 & 99.20 & 99.19 & 99.20 & 99.16 \\
Fold 3 & 99.39 & 99.42 & 99.39 & 99.37 \\
Fold 4 & 99.10 & 99.16 & 99.10 & 99.11 \\
Fold 5 & 98.87 & 98.82 & 98.87 & 98.80 \\
\textbf{Average} & \textbf{99.07} & \textbf{99.09} & \textbf{99.07} & \textbf{99.05} \\
\bottomrule
\end{tabular*}
\end{threeparttable}
\end{table}

The results demonstrate that the KuBERT Transformer consistently delivers high performance across folds, reflecting its robustness in idiom detection. Variability in performance is linked to idioms' contextual diversity and frequency in the dataset. Idioms with broader contextual representations are classified more accurately, while less frequent idioms exhibit slightly lower precision. By leveraging pre-trained Kurdish embeddings and extensive fine-tuning, the model captures nuanced idiomatic meanings, positioning it as a strong candidate for real-world NLP applications involving idiom detection.

\subsection{RCNN Model's Results and Performance}
In each of the five folds, the RCNN model was trained for 50 epochs with a batch size of 16 and a learning rate of 2e-5. Its architecture is designed to include a bidirectional LSTM network, convolution layers, and dropout regularization in an attempt to help learn good contextual information and avoid overfitting. Hereafter, we present the results and metrics obtained for each fold, along with a table summarizing the validation and test metrics for performance evaluation.

\subsubsection{Comparison of the Folds}
Various charts are used to show the performance and trends in performance metrics and validation loss in different folds for the RCNN model. These provide a comparison and hence a great degree of insight into stability, robustness, and the ability to effectively generalize of the model. To visualize the performance comparison across folds, we have two graphs: Figure \ref{fig:RCNN_Validation_Loss} represents the learning curves of validation loss for five folds during 50 epochs of training. This loss starts quite high at approximately 4.0 and rapidly decreases within the first few epochs before remaining constant at around 0.0 beyond epoch 10. All the folds are remarkably similar, with almost identical curves, which is indicative of the model performing consistently well on different splits of the data; it is stable and generalizable. The loss after epoch 5 remains very minimal, with little chance of further reduction; hence, it was reasoned that the model reaches an optimal level quite early, and further training may not result in any significant improvement. The smooth convergence without sharp oscillations suggests that the model is well-tuned, with appropriate hyperparameters, and could have attained similar performance with fewer training epochs.

\begin{figure} [H]
\centering
\includegraphics[width=0.6\linewidth]{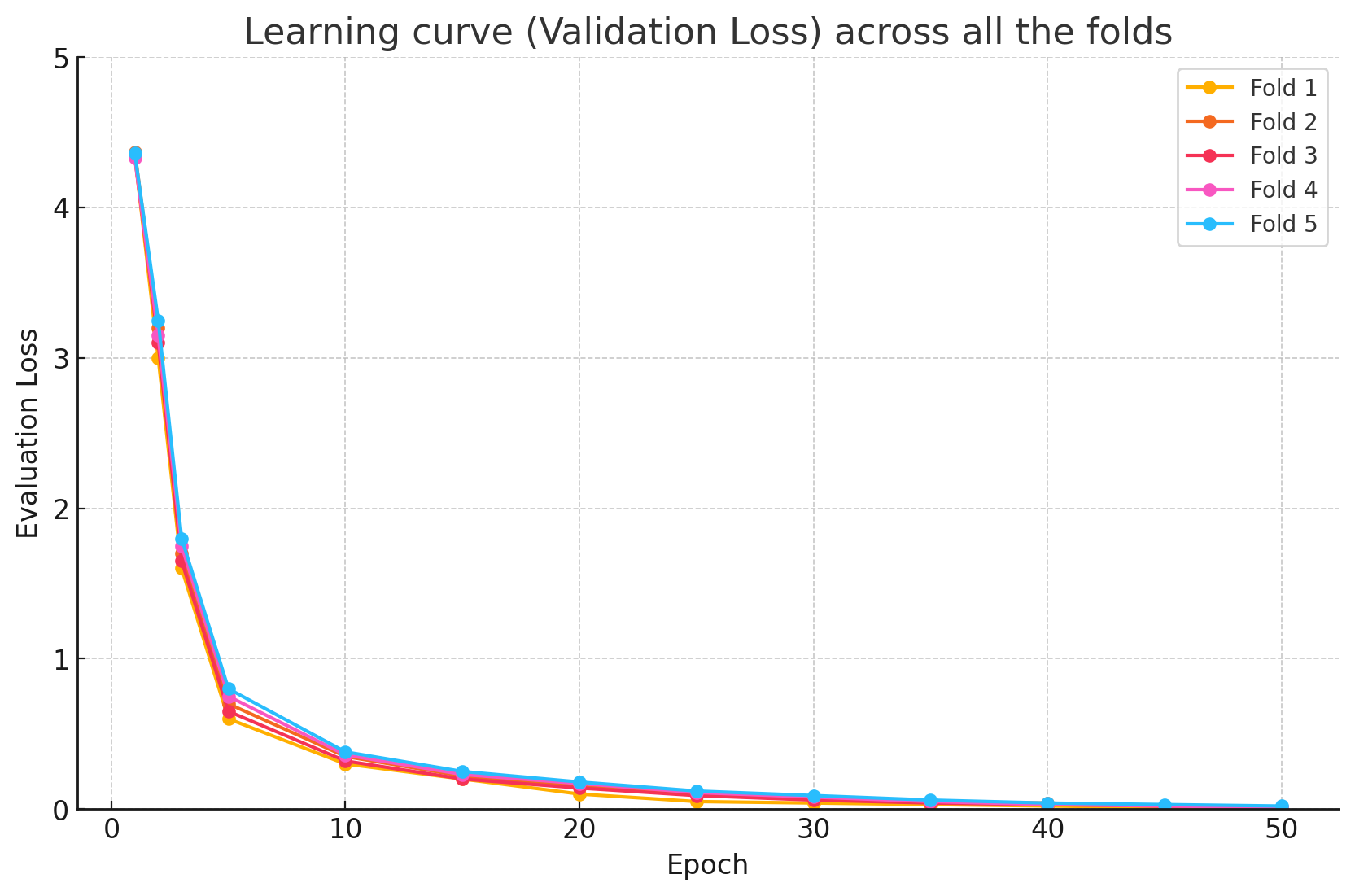}
\caption{RCNN's Learning Curve (Validation Loss) Across all the Folds}
\label{fig:RCNN_Validation_Loss}
\end{figure}

Figure \ref{fig:RCNN_Test_Performance} plots the RCNN model performance over five folds with respect to Accuracy, Precision, Recall, and F1-Score. All metrics follow the same pattern, where performance first improves from Fold 1, peaks in Fold 2, decreases in Fold 3, further lowers in Fold 4, and then partially recovers in Fold 5. In fact, all the metrics are close, which suggests that the model is well-balanced across all measures of performance. It performs best on Fold 2, in which the Accuracy, Precision, Recall, and F1-Score are all remarkably close to 96.6\%. Folds 3 and 4 show some drops, which means the folds are variable, or the model is overfitting. Despite these fluctuations, the model maintains high performance across all folds, with metrics consistently ranging from 96.2\% to 96.7\%.

\begin{figure}[H]
\centering
\includegraphics[width=0.6\linewidth]{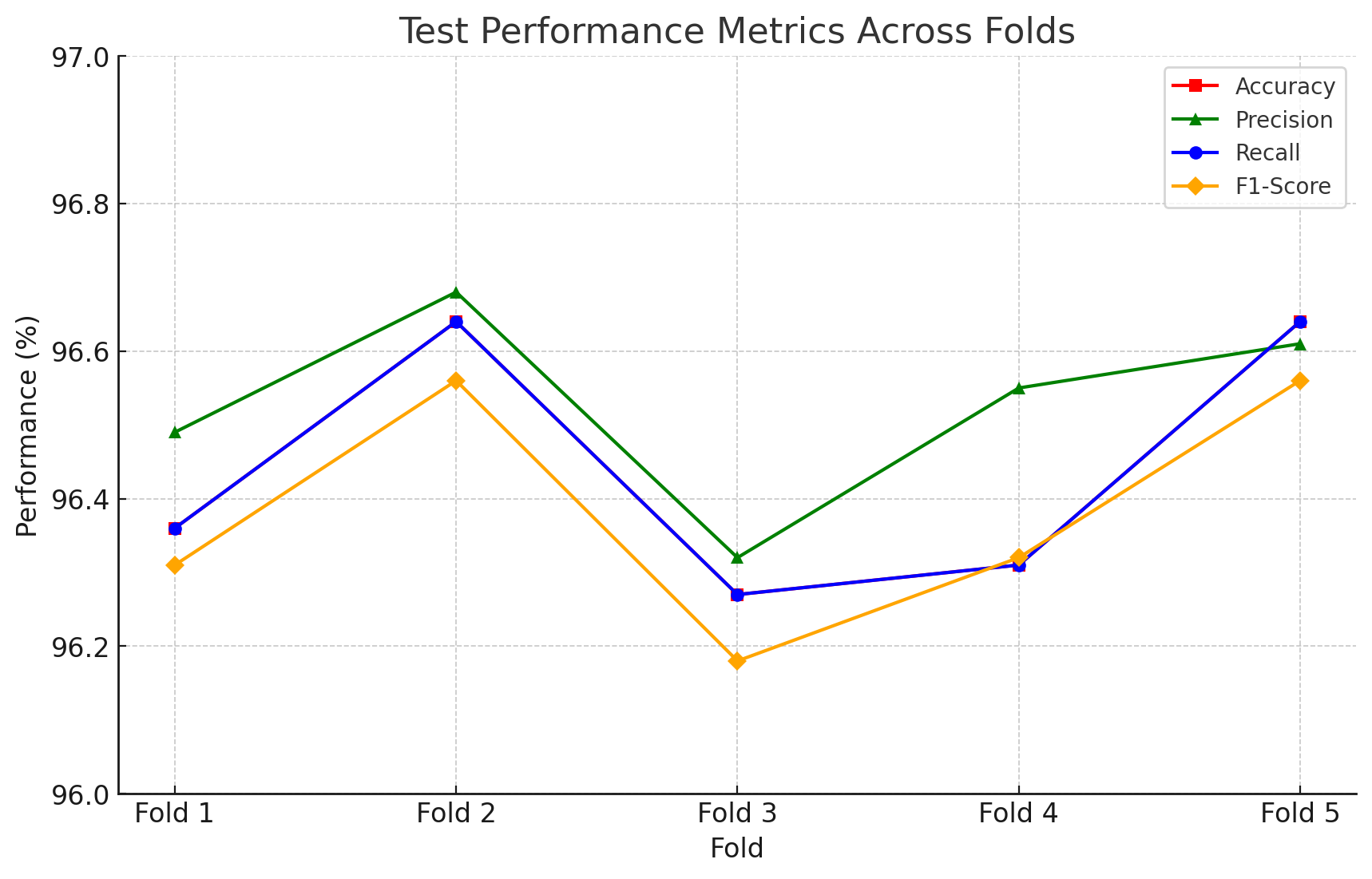}
\caption{RCNN's Test Performance Metrics Across Folds}
\label{fig:RCNN_Test_Performance}
\end{figure}

\subsubsection{Overall Performance}
The overall RCNN model accuracy, averaging across all folds, is 96.45\%, while the precision and recall are 96.53\% and 96.45\%, respectively, with F1-scores of 96.39\%. These measures confirm that the model was effective in performing the task of idiom detection with high generalization capability and minimal variability of performance between different splits of data. The fact that all metrics in this evaluation have high scores shows the model can strike a balance in detecting idioms while keeping false positives low. Precision and Recall are almost identical, meaning the model performs equally well in detecting relevant instances and avoiding misclassifications. The F1-score, balancing Precision and Recall, remains similarly high, further emphasizing the model's robustness and reliability in classification tasks.

Table \ref{table:RCNN_Overall_Performance} summarizes the RCNN model’s performance across all five folds.

\begin{table}[H]
\caption{RCNN's Overall Test Performance Summary Across All Folds}
\label{table:RCNN_Overall_Performance}
\begin{threeparttable}
\begin{tabular*}{\columnwidth}{@{\extracolsep\fill}lcccc@{\extracolsep\fill}}
\toprule
\textbf{Metric} & \textbf{Accuracy (\%)} & \textbf{Precision (\%)} & \textbf{Recall (\%)} & \textbf{F1-Score (\%)} \\
\midrule
Fold 1 & 96.36 & 96.49 & 96.36 & 96.31 \\
Fold 2 & 96.64 & 96.68 & 96.64 & 96.56 \\
Fold 3 & 96.27 & 96.32 & 96.27 & 96.18 \\
Fold 4 & 96.31 & 96.55 & 96.31 & 96.32 \\
Fold 5 & 96.64 & 96.61 & 96.64 & 96.56 \\
\textbf{Average} & \textbf{96.45} & \textbf{96.53} & \textbf{96.45} & \textbf{96.39} \\
\bottomrule
\end{tabular*}
\end{threeparttable}
\end{table}

\subsection{BiLSTM with Attention Model's Results and Performance}
This section presents the performance results of the BiLSTM with Attention model, using the same dataset, KuBERT tokenizer, metrics, and 5-fold cross-validation setup as before. The model was trained for 50 epochs with a batch size of 16 and a learning rate of 2e-5, fine-tuned for optimal learning and stability. Training followed a stratified 5-fold process to maintain class distribution, with validation checks every 5 epochs to track progress and prevent overfitting. The attention mechanism enhanced the model’s focus on key sequence features, improving interpretability and performance. This analysis allows for a direct comparison with the KuBERT model under consistent conditions.

\subsubsection{Comparison of the Folds}
A couple of graphs present important aspects to visualize the model's performance across the folds with respect to performance metrics and validation loss trends. Such a comparison provides insights into the model's stability, robustness, and generalization ability. Figure \ref{fig:BiLSTM_Validation Loss} depicts the validation loss at epochs 10, 20, 30, 40, and 50 across five folds, revealing a consistent downward trend as training progresses. The sharpest decline occurs between epochs 10 and 20, indicating rapid early learning. After epoch 30, the loss reduction slows, showing diminishing returns and the curve stabilizes around epochs 40 to 50. This implies that further training would yield minimal gains or risk of overfitting. The minimal variance between folds indicates robust generalization, with optimal performance likely achieved between 30 and 40 epochs. While additional learning is possible, the marginal improvement may not justify the increased computational cost.

\begin{figure} [H]
    \centering
    \includegraphics[width=0.6\linewidth]{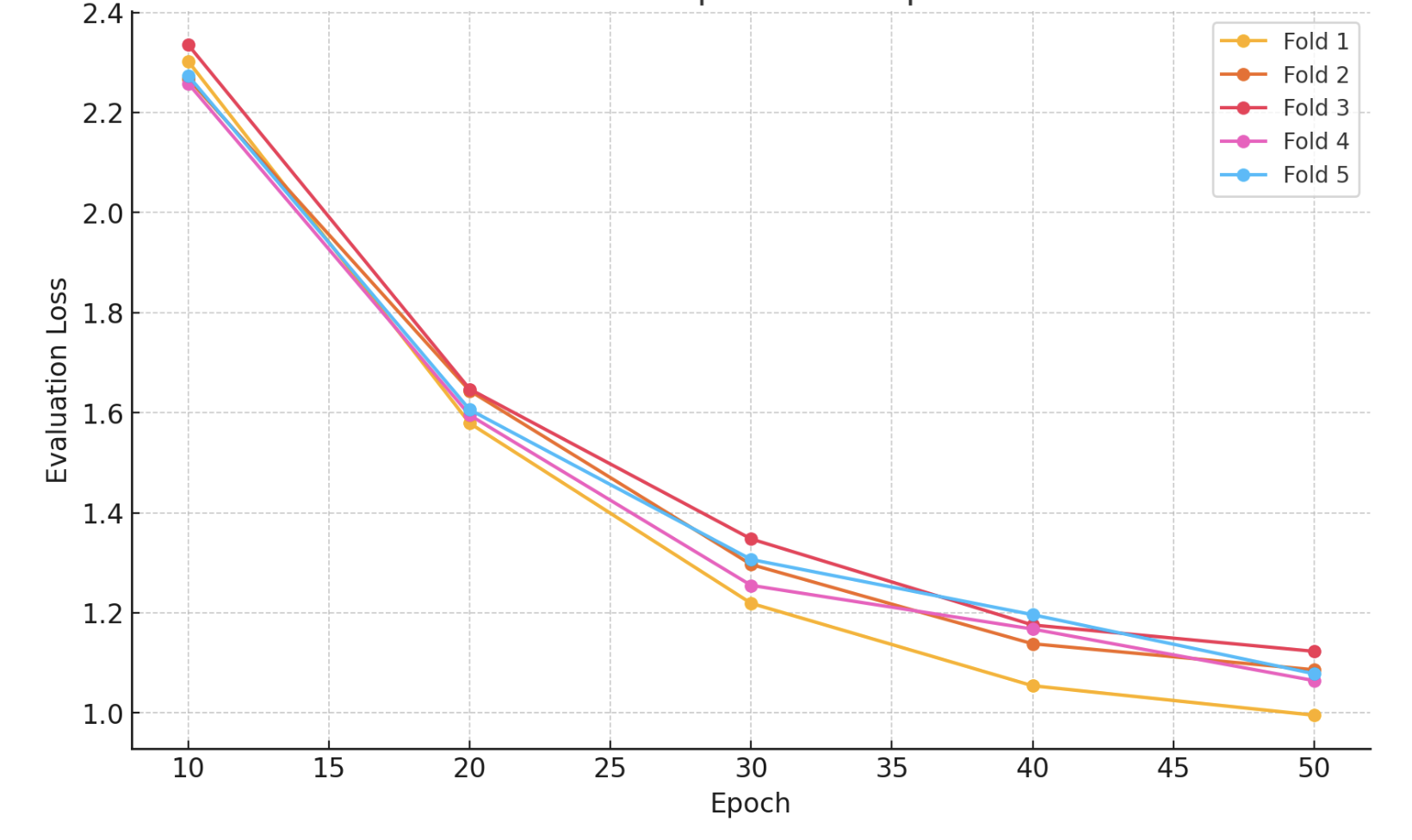}
    \caption{BiLSTM with Attention Learning Curve (Validation Loss) Over 50 Epochs For Each Fold}
    \label{fig:BiLSTM_Validation Loss}
\end{figure}

\subsubsection{Overall Performance}  
The BiLSTM with Attention model consistently performed well across all folds, achieving an accuracy of 80.46\%, with precision, recall, and F1-scores of 80.28\%, 80.46\%, and 79.18\%, respectively. The slightly lower F1-score points to a trade-off between precision and recall, yet the model remains reliable and adaptive across different validation folds. This consistency suggests that the model is robust in handling various scenarios without major performance fluctuations. Its balance between precision and recall makes it well-suited for applications where both false positives and false negatives are costly. Overall, the model’s stability across different contexts shows its reliability in practical applications.

Table \ref{table:biLSTM_Overall_Performance} summarizes the BiLSTM with Attention model’s performance across all five folds.

\begin{table}[H]
\caption{BiLSTM with Attention's Overall Test Performance Summary Across All Folds\label{table:biLSTM_Overall_Performance}}
\begin{threeparttable}
\begin{tabular*}{\columnwidth}{@{\extracolsep\fill}lllll@{\extracolsep\fill}}
\toprule
\textbf{Metric} & \textbf{Accuracy (\%)} & \textbf{Precision (\%)} & \textbf{Recall (\%)} & \textbf{F1-Score (\%)} \\
\midrule
Fold 1 & 81.39 & 81.92 & 81.39 & 80.14 \\
Fold 2 & 80.29 & 79.93 & 80.29 & 79.01 \\
Fold 3 & 78.69 & 79.42 & 78.69 & 77.41 \\
Fold 4 & 79.21 & 78.22 & 79.21 & 77.87 \\
Fold 5 & 82.70 & 81.94 & 82.70 & 81.46 \\
\midrule
\textbf{Average} & \textbf{80.46} & \textbf{80.28} & \textbf{80.46} & \textbf{79.18} \\
\bottomrule
\end{tabular*}
\end{threeparttable}
\end{table}

In summary, the BiLSTM with Attention model delivered strong, consistent results across all five folds in cross-validation. With an overall accuracy of 80.46\%, along with precision, recall, and F1-scores averaging 80.28\%, 80.46\%, and 79.18\%, the model demonstrates its effectiveness in idiom detection. The evaluation shows the model maintains a balanced approach, with minimal performance fluctuations between folds. While there is a slight dip in Folds 2 and 3, the model recovers and maintains stability, particularly in precision, indicating its robustness in reducing false positives. The learning curve suggests optimal performance is achieved between 30 and 40 epochs, with diminishing returns beyond that point, showing effective convergence.

\subsection{Comparative Analysis of the Models' Performance}

This section compares the performance of KuBERT Transformer, BiLSTM with Attention, and RCNN on the same dataset using cross-validation. Key performance metrics include accuracy, precision, recall, F1-score, training time, and computational efficiency.

\subsubsection{Performance Metrics}

 The KuBERT Transformer consistently outperformed the other two models in accuracy, precision, recall, and F1-score, averaging close to 99\%. The RCNN model also showed steady performance, with metrics averaging around 96\%, while the BiLSTM with Attention lagged, averaging around 80\% across all metrics. Table \ref{table:comparison_overall_performance} summarizes the overall performance of each model across all folds.

\begin{table}[H]
\caption{Overall Performance Comparison Across All Models\label{table:comparison_overall_performance}}
\begin{threeparttable}
\begin{tabular*}{\columnwidth}{@{\extracolsep\fill}lllll@{\extracolsep\fill}}
\toprule
Model & Accuracy (\%) & Precision (\%) & Recall (\%) & F1-Score (\%) \\
\midrule
KuBERT Transformer & 99.07 & 99.09 & 99.07 & 99.05 \\
RCNN & 96.45 & 96.53 & 96.45 & 96.39 \\
BiLSTM with Attention & 80.46 & 80.28 & 80.46 & 79.18 \\
\bottomrule
\end{tabular*}
\end{threeparttable}
\end{table}

Figure \ref{fig:performance_comparison_graph} highlights the KuBERT Transformer as the top-performing model, achieving the highest metrics across all folds with balanced precision and recall, excelling in idiom detection tasks. The RCNN, while not surpassing the Transformer, delivered consistent and solid results with minimal fluctuations. In contrast, the BiLSTM with Attention struggled significantly, showing lower precision and recall, which led to notable misclassifications and a large performance gap compared to the other models.

\begin{figure}[H]
\centering
\includegraphics[width=0.6\linewidth]{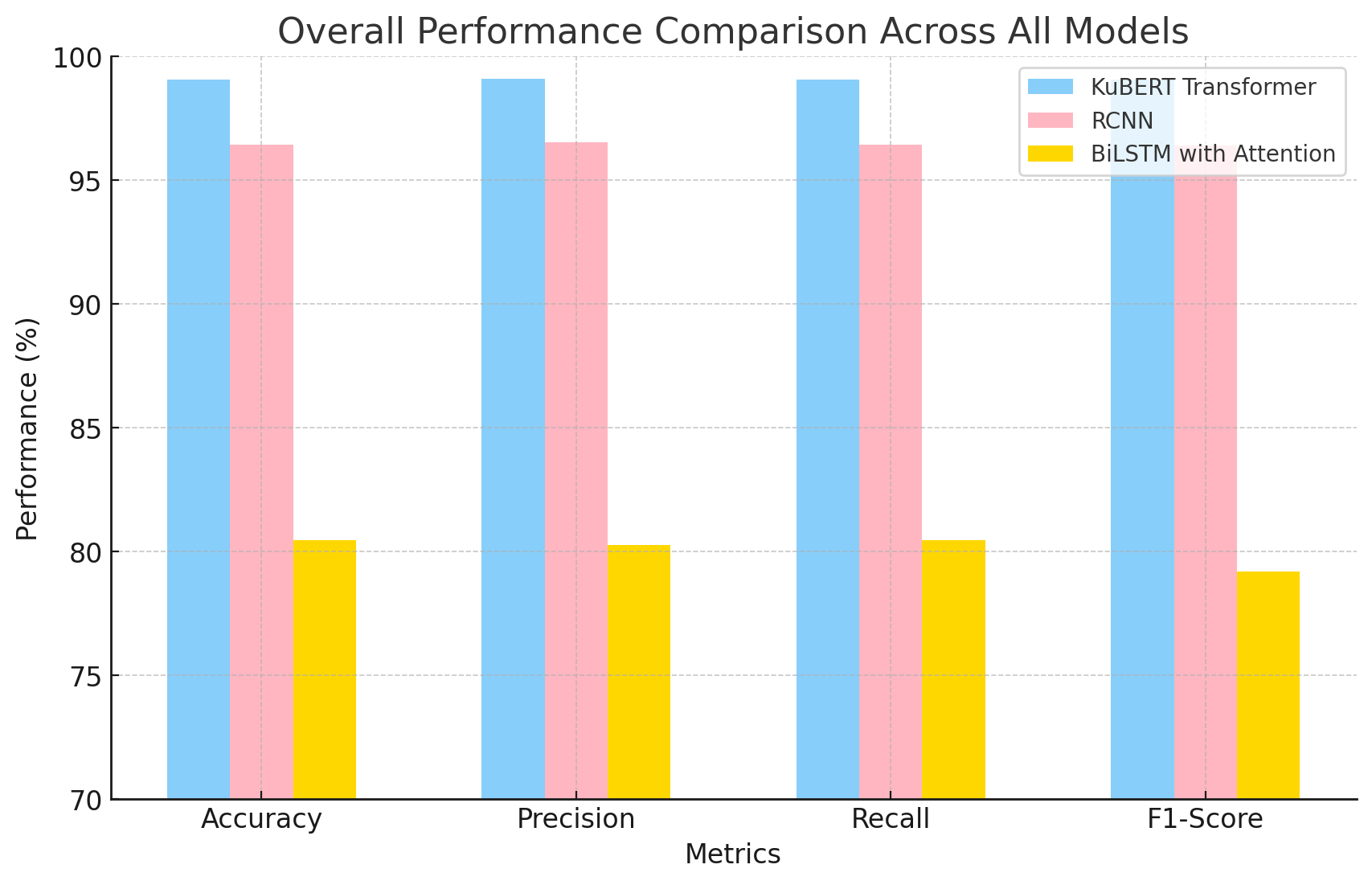}
\caption{Performance Metrics Comparison Across Models}
\label{fig:performance_comparison_graph}
\end{figure}

\subsubsection{Performance Stability Across Folds}

Figure \ref{fig:comparison_test_metrics} shows the test metrics for each model across all folds. The KuBERT Transformer displayed the most consistent performance, with minimal fluctuations in metrics, indicating strong stability and generalization. The RCNN model also maintained stable performance, though some minor dips were observed in a few folds. However, the BiLSTM with Attention showed greater variability, particularly in Folds 2 and 3, where accuracy and recall noticeably dropped.

\begin{figure}[H]
\centering
\includegraphics[width=0.6\linewidth]{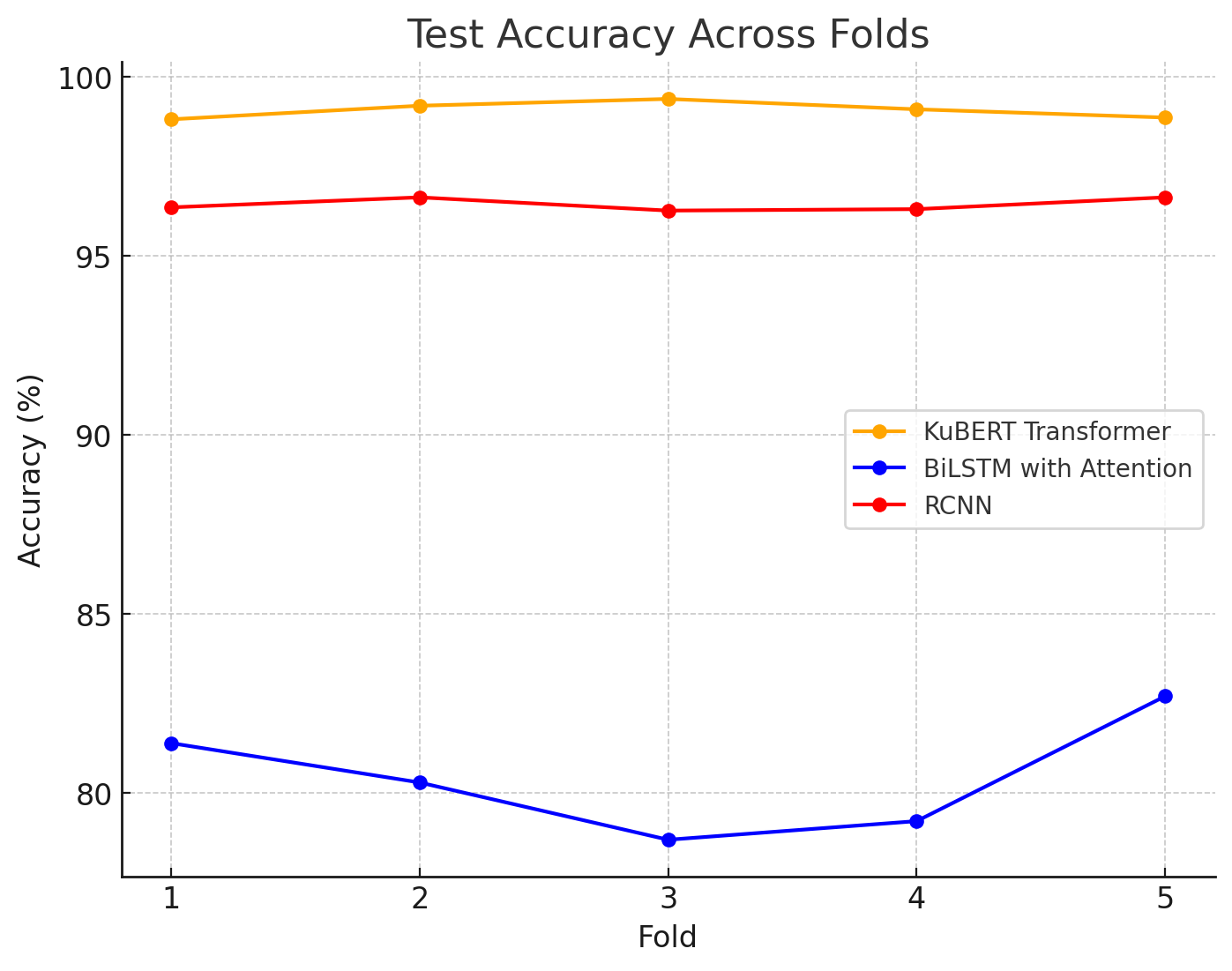}
\caption{Comparison Test Accuracy Across Folds and Models}
\label{fig:comparison_test_metrics}
\end{figure}

\subsubsection{Learning Curve Analysis}

Figure \ref{fig:LearningCurveComparison} illustrates the learning curves of training and validation losses for the three models. The KuBERT Transformer demonstrates rapid convergence within 15 epochs, achieving nearly zero loss values and small differences between training and validation losses, indicating excellent generalization and efficiency. Conversely, the BiLSTM with Attention requires up to 50 epochs for reasonable convergence, with most improvements occurring in the first 20 epochs. However, a noticeable gap between its training and validation losses raises concerns about overfitting. The RCNN shows a steep decline in loss within the initial epochs, stabilizing around epoch 30, offering more balance compared to BiLSTM. Among the models, the KuBERT Transformer stands out for its speed and stability, followed by the RCNN, while the BiLSTM lags with slower convergence and potential overfitting issues.

\begin{figure}[H]
\centering
\includegraphics[width=0.6\linewidth]{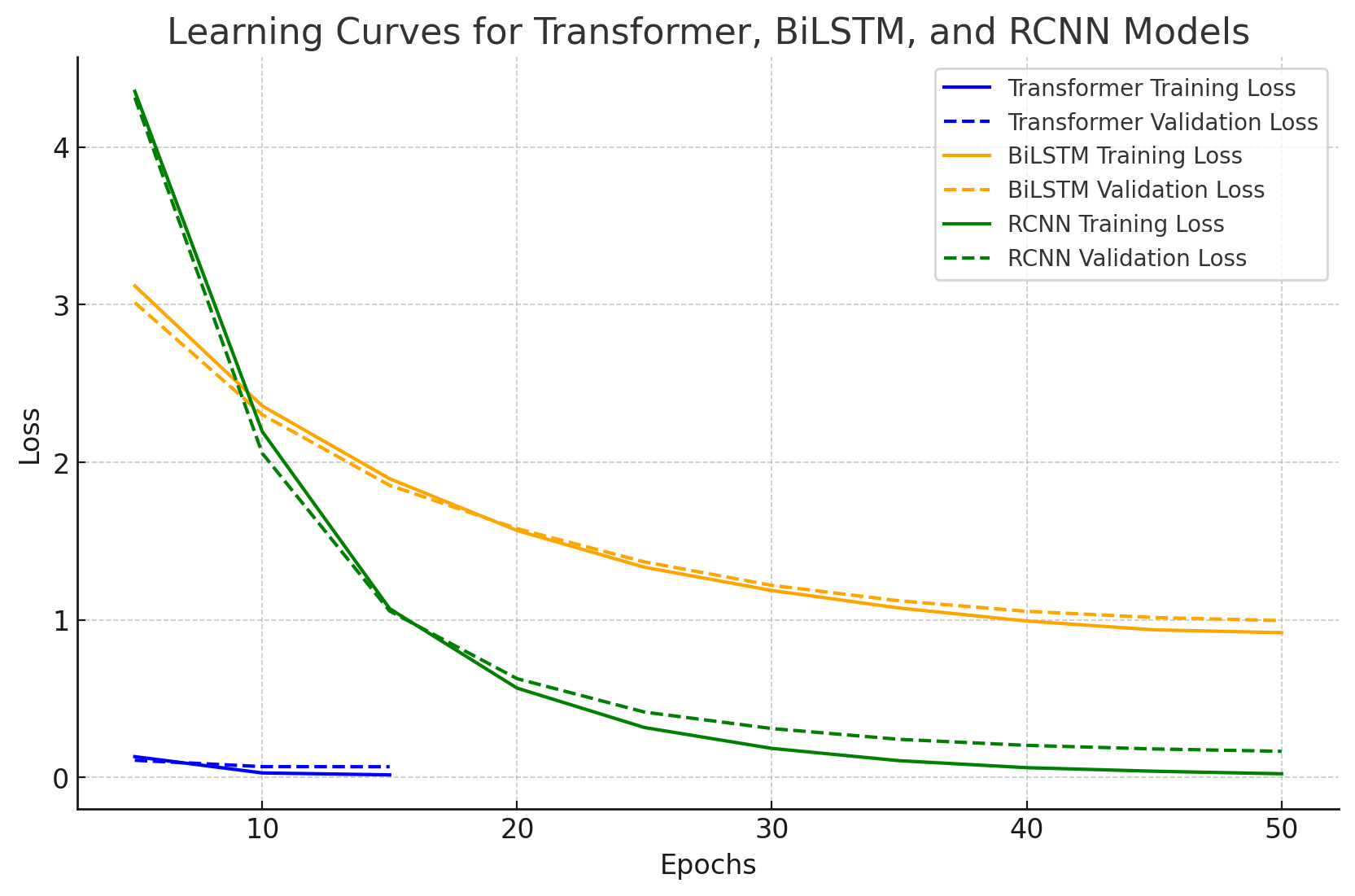}
\caption{Learning Curves for Transformer, BiLSTM, and RCNN Models}
\label{fig:LearningCurveComparison}
\end{figure}

\subsubsection{Training Duration and Computational Efficiency}
Table \ref{table:comparison_training_duration} compares training durations for each model. The BiLSTM with Attention required the longest time, averaging 10 hours per fold, but delivered the lowest performance, making it the least efficient. The RCNN was far more efficient, training in about 1 hour per fold with competitive performance. The KuBERT Transformer, while requiring 8.5 hours per fold and higher computational resources, achieved the best performance, justifying its longer training time.

\begin{table}[H]
\caption{Training Duration Comparison Across Models\label{table:comparison_training_duration}}
\begin{threeparttable}
\begin{tabular*}{\columnwidth}{@{\extracolsep\fill}lll@{\extracolsep\fill}}
\toprule
Model & Avg. Training Time/Fold & Total Elapsed Time \\
\midrule
KuBERT Transformer & 8 hrs 24 mins & 42 hrs \\
BiLSTM with Attention & 10 hrs 41 mins & 53 hrs 25 mins \\
RCNN & 1 hr 3 mins & 5 hrs 15 mins \\
\bottomrule
\end{tabular*}
\end{threeparttable}
\end{table}

\subsection{Discussion}
The comparative analysis of the KuBERT Transformer, BiLSTM with Attention, and RCNN models reveals their strengths and limitations in idiom detection. The KuBERT Transformer outperformed the others, achieving 99\% accuracy by leveraging pre-trained Kurdish embeddings and self-attention mechanisms to capture contextual nuances. The RCNN followed with a solid 96.5\% accuracy, offering a balance between efficiency and performance, while the BiLSTM with Attention lagged at around 80\%, struggling with complex idioms and showing high variability.

The KuBERT Transformer converged quickly within 15 epochs, with minimal divergence between training and validation losses, making it highly stable and reliable for real-world applications. Although it requires significant computational resources (8.5 hours per fold), its superior performance justifies this cost. The RCNN, which trains in about 1 hour per fold, is a more efficient option for resource-limited scenarios. In contrast, the BiLSTM's slower convergence and higher risk of overfitting, combined with its long training time (10 hours per fold), make it less suitable.

Despite KuBERT's strong performance, challenges remain, such as high computational costs and difficulties with rare or context-specific idioms. Future work could explore optimization techniques, hybrid models combining RCNN and Transformer, and dataset expansion for better generalization. This study emphasizes the need to select models based on task-specific requirements, balancing accuracy, efficiency, and complexity.

\section{Conclusion and Future Work}
This thesis addresses the gap in NLP for Sorani Kurdish by focusing on automated idiom detection, a challenging task due to idioms' non-literal meanings and varied usage across contexts. With limited prior research in Kurdish idiom detection, misinterpretations often arise in sentiment analysis, machine translation, and educational applications. This work aimed to bridge this gap by creating a dataset of 101 Kurdish idioms embedded in 10,580 sentences, carefully designed to represent diverse contexts and grammatical structures. Kurdish linguists reviewed these examples to ensure reliability and provide a robust foundation for future Kurdish NLP research. Using this dataset, three models were developed and evaluated on idiom detection: KuBERT-based Transformer Sequence Classification, a BiLSTM with Attention, and an RCNN. Stratified k-fold cross-validation was applied to address class imbalance and ensure comprehensive performance evaluation.

The KuBERT Transformer model outperformed the others, achieving 99\% accuracy by leveraging pre-trained Kurdish embeddings and an architecture adept at capturing subtle contexts. The RCNN model performed well with 96.5\% accuracy but was less stable across folds, while the BiLSTM with Attention lagged at 80\%, struggling with idiomatic complexity. Despite the promising results, limitations included the small dataset size, potentially affecting generalization, and the computational demands of training KuBERT. Nonetheless, this study makes a significant contribution to Kurdish NLP, providing a robust model and dataset, highlighting the potential of Transformer-based models for idiom detection, and advancing research in this under-resourced language.

This study lays a strong foundation for automated idiom detection in Sorani Kurdish while highlighting areas for further development to advance Kurdish NLP. Expanding the dataset to include more idioms could enhance model generalizability for broader applications. Exploring hybrid models that combine the RCNN's efficiency with the Transformer's contextual depth may yield improved results. Implementing span-based extraction approaches to identify idiom positions within sentences could enhance precision, particularly for idioms in longer texts. Additionally, integrating these models into real-time applications such as chatbots, translation systems, or educational tools could demonstrate their practical value and contribute to cultural preservation.

\section*{Acknowledgment}
We are deeply grateful to those who assisted us in the data collection and throughout this study. We acknowledge and appreciate the assistance of Dazgay Mukryani, particularly Kak Mohyadin, for providing us with Sorani idioms. We are also grateful to Mamosta Abdulwahab Shexani for sharing valuable books on Kurdish idioms with us. Thank you to Mamosta Bahzad for his linguistic expertise and for carefully reviewing the dataset with embedded idioms to ensure accuracy. Additionally, we acknowledge the support of Dr. Azad, Dr. Mzgeen, and their students at Salahaddin University for their invaluable assistance in the proper text preparation with the idioms embedded. We express our heartfelt gratitude to Mr. Abdulhady Abas, who suggested KuBERT and gave us insightful ideas on its applications. Our appreciation also goes to Mr. Goran Khursheed Faisal for his valuable technical support where needed. We remain appreciative to all those who assisted us, but their names have not been mentioned due to their preference.

\bibliographystyle{lrec}
\bibliography{references}

\end{document}